\documentclass{article}
\usepackage[dvips]{graphicx}

\usepackage{times}
\usepackage{rotating}
\usepackage{epsfig}
\usepackage{graphicx}
\usepackage{amsmath}
\usepackage{amsthm}
\usepackage{amssymb}
\usepackage{multirow}
\usepackage{tabularx}
\usepackage{booktabs}
\usepackage{url,color}

\newcommand\eqdef{\overset{\triangle}{=}}

\newcommand{\BEAS}{\begin{eqnarray*}}
\newcommand{\EEAS}{\end{eqnarray*}}
\newcommand{\BEA}{\begin{eqnarray}}
\newcommand{\EEA}{\end{eqnarray}}
\newcommand{\BEQ}{\begin{equation}}
\newcommand{\EEQ}{\end{equation}}
\newcommand{\BIT}{\begin{itemize}}
\newcommand{\EIT}{\end{itemize}}
\newcommand{\BNUM}{\begin{enumerate}}
\newcommand{\ENUM}{\end{enumerate}}
\newcommand{\BA}{\begin{array}}
\newcommand{\EA}{\end{array}}

\newcommand{\rb}{\mathbb{R}}

\newcommand{\vy}{\mathbf{y}}

\newcommand{\mysec}[1]{Section~\ref{sec:#1}}
\newcommand{\eq}[1]{Eq.~(\ref{eq:#1})}
\newcommand{\myfig}[1]{Figure~\ref{fig:#1}}

\theoremstyle{definition}

\newcommand{\norm}[1]{\|{#1}\|}

\newcommand{\half}{\tfrac{1}{2}}

\DeclareMathOperator*{\argmax}{argmax}

\begin{document}

\title{Convex Optimization for Parallel Energy Minimization}
\date{\today\footnote{Based on work  originally submitted on 14 November, 2014.}}
\author{K. S. Sesh Kumar\\
INRIA-Sierra project-team\\ 
D{\'e}partement d{'}Informatique \\
de l{'}Ecole Normale Sup{\'e}rieure\\
Paris, France\\
\texttt{sesh-kumar.karri@inria.fr}\\
\and {\'A}lvaro Barbero\\
Instituto de Ingenier{\'i}a del Conocimiento \\ and Universidad Aut{\'o}noma de Madrid\\
Madrid, Spain\\
\texttt{alvaro.barbero@uam.es}
\and Stefanie Jegelka \\
Department of EECS\\
Massachusetts Institute of Technology\\
Cambridge, USA\\
\texttt{stefje@csail.mit.edu}
\and  Suvrit Sra\\
Department of EECS\\
Massachusetts Institute of Technology\\
Cambridge, USA\\
\texttt{suvrit@mit.edu}
\and Francis Bach\\
INRIA-Sierra project-team\\ 
D{\'e}partement d{'}Informatique \\
de l{'}Ecole Normale Sup{\'e}rieure\\
Paris, France\\
\texttt{francis.bach@inria.fr}
}\maketitle

\begin{abstract}
Energy minimization has been an intensely studied core problem in computer vision. With growing image sizes (2D and 3D), it is now highly desirable to run energy minimization algorithms in parallel. But many existing algorithms, in particular, some efficient combinatorial algorithms, are difficult to parallelize. By exploiting results from convex and submodular theory, we reformulate the quadratic energy minimization problem as a total variation denoising problem, which, when viewed geometrically, enables the use of projection and reflection based convex methods. The resulting min-cut algorithm (and code) is conceptually very simple, and solves a sequence of TV denoising problems. We perform an extensive empirical evaluation comparing state-of-the-art combinatorial algorithms and convex optimization techniques. On small problems the iterative convex methods match the combinatorial max-flow algorithms, while on larger problems they offer other flexibility and important gains: (a) their memory footprint is small;
(b) their straightforward parallelizability fits multi-core platforms; (c) they can easily be warm-started; and (d) they quickly reach approximately good solutions, thereby enabling faster ``inexact'' solutions. A key consequence of our approach based on submodularity and convexity is that it is allows to combine \emph{any arbitrary combinatorial or convex methods as subroutines}, which allows one to obtain hybrid combinatorial and convex optimization algorithms that benefit from the strengths of both.
\end{abstract}

\section{Introduction}

Energy minimization has become a key element in many low- to mid-level tasks in computer vision, such as segmentation or stereo correspondence (see \cite{blake11book} for a survey). For many, frequently occurring such minimization problems, graph cut techniques have emerged as generic, very efficient tools that provide global optima for submodular quadratic penalties, and extend to several higher-order and non-submodular potentials too. However, when applying widely used graph-cut code (e.g.,~\cite{boykov2001fast}) to huge problems in 3D or video, running time and memory usage become problematic. Ideally, we would wish to have algorithmic flexibility to decompose the problem into (almost) arbitrary subproblems that can be solved in parallel, and that adapt to new architectures such as GPU clusters.
These latter needs can be met through convex optimization. 

In this paper, we explore methods that allow combining convex and combinatorial optimization, and thereby offer a way to parallelize recent successful combinatorial methods \cite{boykov04,Chandran:2009,ibfs}. Our algorithms can run on large datasets while using only limited memory, and are flexible enough to be ported to different hardware architectures.

We begin by rewriting the energy minimization problem as a convex optimization problem.
More precisely, we consider the equivalent reformulation of the min-cut problem as the minimization of a unary term $w^\top x$ plus the total variation $f(x)$ associated to a graph, over the vertices of the hypercube $x \in \{0,1\}^n$. 
The straightforward convex relaxation 
replaces the set of vertices $\{0,1\}^n$ of the hypercube with the full hypercube $[0,1]^n$. This relaxation is tight, a fact that has been exploited widely. Moreover, typical graph structures that occur in low-level vision problems naturally decompose as a superposition of chains, suggesting the use of iterative convex optimization methods~\cite{komodakis2011mrf,savchynskyy11}.
However, the lack of smoothness (in the primal and dual problems) poses some difficulties in solving the problem efficiently.

Hence, we instead consider a reformulation through total variation denoising outlined by~\cite{hochbaum01,treesubmod} for general submodular functions, which we use for cuts in this paper. The gist of this approach is to replace the non-smooth relaxation by the 
total variation (TV) denoising problem $\min_{x \in \rb^n} \frac{1}{2} \| x - w\|^2 + f(x)$, from which one may obtain an optimal solution of the energy minimization problem by thresholding. The benefit of this formulation is its smooth dual problem, which has a natural geometric interpretation: it reduces to computing the distance between two convex sets. Moreover, via the equivalence between energy minimization and projections we obtain fast projection subroutines, which in turn enable the use of classical, popular projection methods~\cite{treesubmod}.
 
While we focus on cuts in this paper, our framework extends to all higher-order submodular potentials, non-submodular quadratic potentials, and to multi-label problems (see Section~\ref{sec:extensions} for more details).
We make the following contributions:
\begin{itemize}
\setlength{\itemsep}{-1pt}
\item[--] We study algorithms that allow to easily combine convex and combinatorial optimization (e.g., graph cut, total variation, algorithms for higher order submodular potentials), and to parallelize existing fast energy minimization methods for submodular potentials.
In fact, this approach applies to arbitrary sums of submodular functions, each defined on all or a subset of the variables.
\item[--] As a case study, we specialize the optimization framework of~\cite{treesubmod} to the submodular energy functions most commonly used in computer vision, such as quadratic functions with a 2D or 3D grid structure with various connectivities (Section~\ref{sec:2D3D}).
The resulting min-cut algorithm (and code) is conceptually very simple 
and may thus be easily used and modified.

\item[--] We perform an extensive empirical evaluation of serial and parallel, combinatorial and convex min-cut codes on benchmarks~\cite{verma12,kappes13}. We observe that iterative techniques based on convex optimization match combinatorial max-flow algorithms on small problems, and on larger problems they offer complementary flexibility: (a) they have a reduced memory footprint and can tackle many problems where traditional methods run out of memory, (b) they parallelize easily on multi-core platforms, (c) they can be efficiently warm-started, (d) they quickly reach approximately good solutions, thereby enabling faster ``inexact'' solutions. These observations suggest to marry convex and combinatorial methods to attain  ``the best of both worlds''.

\end{itemize}

\section{Convex optimization for graph cuts}
In this paper, we focus on energy minimization problems with pairwise potentials,
\begin{equation}
  \label{eq:1}
  E(x) = - \sum\nolimits_{i=1}^n w_i x_i + \sum\nolimits_{i,j=1}^n \psi_{ij}(x_i,x_j),
\end{equation}
where the variables $x_i$ take values in a set of discrete labels. For simplicity, we here focus on binary labels, $x_i \in \{0,1\}$. The algorithms we discuss may be generalized to multiple labels via move-making strategies \cite{boykov2001fast}. Moreover, we assume the pairwise potentials to be submodular, i.e., $\psi_{ij}(0,1) + \psi_{ij}(1,0) \geq \psi_{ij}(0,0) + \psi_{ij}(1,1)$. (One may extend to non-submodular potentials via roof duality \cite{rother2007optimizing}). It is well known that all such submodular energy functions may be written as graph cut functions with nonnegative ``edge weights'' $a_{ij}$, up to a constant \cite{picard75}:
\begin{equation}
  \label{eq:4}
  E(x) = - \sum\nolimits_{i=1}^n w_i x_i + \sum\nolimits_{i,j=1}^n a_{ij} | x_i - x_j | + \mathrm{const.}
\end{equation}
This function consists of two parts: (1) a sum of unary potentials $-\sum_{i=1}^n w_i x_i  =  - w^\top x$; and (2) the sum of pairwise potentials, which is equivalent to a weighted graph cut between the set of indices $i$ in $\{1,\dots,n\}$ for which $x_i=1$, and its complement. For $x \in \mathbb{R}^n$, this sum is the \emph{total variation function} $f(x) \eqdef \sum_{i,j=1}^n a_{ij} | x_i - x_j |$. Note that this is a case of anisotropic weighted total variation. Since the weights $a_{ij}$ are non-negative, the function $f$ is convex.
We refer to the graph cut problem as the \emph{discrete problem}:
\BEQ
\label{eq:D}
\tag{D}
\min_{x \in \{0,1\}^n} f(x) - w^\top x.
\EEQ

\subsection{Convex reformulation}
We obtain a relaxation to the combinatorial problem in \eq{D} by replacing $\{0,1\}^n$ by its convex hull $[0,1]^n$:
\BEQ
\label{eq:C}
\tag{C}
\min\nolimits_{x \in [0,1]^n} f(x) - w^\top x.
\EEQ
We refer to \eq{C} as the \emph{continuous problem}. This relaxation is \emph{exact}: 
since the continuous convex problem in \eqref{eq:C} is a minimization problem over a larger set than the discrete problem, its minimal value has to be lower than \eqref{eq:D}. However, as a consequence of properties of the total variation and its relation to submodular graph cut functions (see, e.g.,~\cite[Sec.~3.3]{fot_submod} or~\cite{hochbaum01,chambolle2009total} for a proof dedicated to cut functions), the 
two optimal values are equal and a solution to \eqref{eq:D} may be obtained from a solution $x \in [0,1]^n$ of \eqref{eq:C} by looking at all ``level sets'' of $x$, that is by rounding the values of $x$ to zero or one by thresholding at a given level in $ [0,1]$ (there are at most $n$ possible thresholds, which can be obtained by first sorting the components of $x$).

\subsection{Convex duality}
The total variation $f$ is convex and absolutely homogeneous, that is, for any $x \in \rb^n$ and $\lambda \in \rb$, $f( \lambda x) = |\lambda|  f(x)$. For all such functions, there exists a centrally symmetric convex body $K \subset \rb^n$ such that for all $x \in \rb^n$~\cite[\S13]{rockafellar97},
$$
f(x) = \max\nolimits_{y \in K}\ y^\top x.
$$
Note that when $f(x)$ happens to be equal to zero only for $x=0$, then $f$ is a norm and the set $K$ is simply the unit ball of the dual norm.

Since $f$ is piecewise affine, the set $K$ is a \emph{polytope} (i.e., the convex hull of finitely many points). The set $K$ may be described precisely for general submodular functions and is usually referred to as the \emph{base polytope} of the submodular function~\cite{fujishige2005submodular,treesubmod,fot_submod}.

Using Fenchel duality, we arrive at the following dual problem to \eqref{eq:C} \cite{fujishige2005submodular,fot_submod}:
\BEA
\nonumber 
\!\!\!\!\!\!\!\!\! \min_{x \in [0,1]^n}  f(x)  - w^\top x
& \!\!= \!\!& \min_{x \in [0,1]^n}  \max_{y \in K} y^\top x  - w^\top x \\
\nonumber & \!\!= \!\!&  \max_{y \in K} \min_{x \in [0,1]^n} x^\top ( y - w)  \\
\label{eq:C-dual}& \!\!= \!\! & \max_{y \in K} \sum_{i=1}^n \min\{ y_i - w_i, 0\}.
\EEA

This dual problem allows us to obtain \emph{certificates of optimality}: given a pair $(x,y) \in [0,1]^n \times K$, the quantity
$$
{\rm gap}(x,y) := f(x) - w^\top x - \sum_{i=1}^n \min\{ y_i - w_i, 0\}
$$
is always non-negative, and equal to zero if and only if $x$ is optimal for \eqref{eq:C} and $y$ is optimal for the dual problem \eq{C-dual}. This duality relation is essential to certify that a given solution is optimal (and corresponds to the traditional min-cut/max-flow duality).

While the cut problem is now reformulated as a convex optimization problem, it is still hard to minimize because neither the primal nor the dual are smooth, and thus iterative methods are typically slow (see detailed comparisons in~\cite{treesubmod}). We now reformulate the problem so that the dual problem becomes smooth and potentially easier to optimize.

\subsection{Equivalence to total-variation denoising}
Following~\cite{fujishige2005submodular,fot_submod,treesubmod,nagano11}, we consider the \emph{total variation denoising problem}:
\BEQ
\label{eq:TV}
\tag{TV}
\min\nolimits_{x \in \rb^n} f(x) + \frac{1}{2} \| x - w\|^2.
\EEQ
By expanding $\frac{1}{2} \| x - w\|^2$ into $\frac{1}{2} \| x\|^2 - w^\top x + \frac{1}{2} \|  w\|^2$, we see that going from the continuous problem \eqref{eq:C} to \eqref{eq:TV} means replacing the constraint $x \in [0,1]^n$ by the penalty $\frac{1}{2} \| x \|^2 = \frac{1}{2} \sum_{i=1}^n x_i^2$. This has a number of important consequences:
(1) It makes the optimization problem strongly convex and thus the dual problem will be smooth; see \eq{TV-dual}.
(2)
A solution to \eqref{eq:D} and hence \eqref{eq:C} may be obtained by \emph{thresholding} the unique solution $x$ of \eqref{eq:TV} at zero, that is, by defining $\hat{x}_i=1$ if $x_i>0$ and $\hat{x}_i =0$ otherwise. This is usually not true for arbitrary convex functions $f$ (even absolutely homogeneous) and is a direct consequence of submodularity.

Importantly, we \emph{need not} solve the TV problem~\eqref{eq:TV} exactly to obtain a solution to~\eqref{eq:C}, we only need to know which of the components are positive (resp.\  negative). 

Analogously to \eq{C-dual}, we may derive the dual to \eqref{eq:TV}:
\BEA
\nonumber 
\!\!\!\!\! \min_{x \in \rb^n}  f(x)  + \frac{1}{2} \| w - x\|^2  
&\!\! \!\!= \!\!\!\! & \min_{x \in \rb^n}  \max_{y \in K} y^\top x  + \frac{1}{2} \| w - x\|^2 \\
\nonumber & \!\!\!\! = \!\! \!\!&  \max_{y \in K} \min_{x \in \rb^n} y^\top x  + \frac{1}{2} \| w - x\|^2   \\
\label{eq:TV-dual}& \!\!\!\! = \!\!  \!\! & \max_{y \in K} \frac{1}{2} \| w\|^2 - \frac{1}{2} \| y -w\|^2.
\EEA
The primal and dual solutions $x$ and $y$ have a simple correspondence $x = w - y$, as can be seen from the dual derivation.
We have now obtained a dual problem which may be simply interpreted as the \emph{orthogonal projection} of the vector $w$ onto the polytope $K$.
Importantly, a slight additional argumentation shows that the discrete energy minimization problem (D), the total variation relaxation (TV), and the projection onto $K$ are equivalent. That means, a fast subroutine for one of the problems implies a fast subroutine for the other two.

While the reformulation is intuitive, orthogonal projections onto the polytope $K$ are not fast to compute in general.  
Many special cases, however, are fast, including graphs that have fast cut subroutines (which may be combinatorial).

\section{Decomposition of graphs}
We assume that the total variation $f$ (and equivalently the energy function) may be decomposed into a sum $\sum_{j=1}^r f_j(x)$ of $r$ total variation functions $f_j$. 
Each function $f_j$ can be defined on 2D grids, trees or chains. The only criterion for the decomposition is to efficiently perform orthogonal projections onto the polytope $K_j$ corresponding to
$f_j$. In fact, $f_j$ does not need to have full support.

While other decompositions are possible \cite{kolmogorov2006convergent,sontag09}, in this paper, we focus on chains  to illustrate our ideas; we note that 1D-TV (on chains) can be solved very efficiently--see Section~\ref{sec:1DTV}. Note that the same idea works for decompositions into trees, stars, small ``cubes'' and 2D sheets.

\subsection{2D and 3D grids}
\label{sec:2D3D}

\begin{figure}[tbp]
  \centering
  \includegraphics[width=0.28\textwidth]{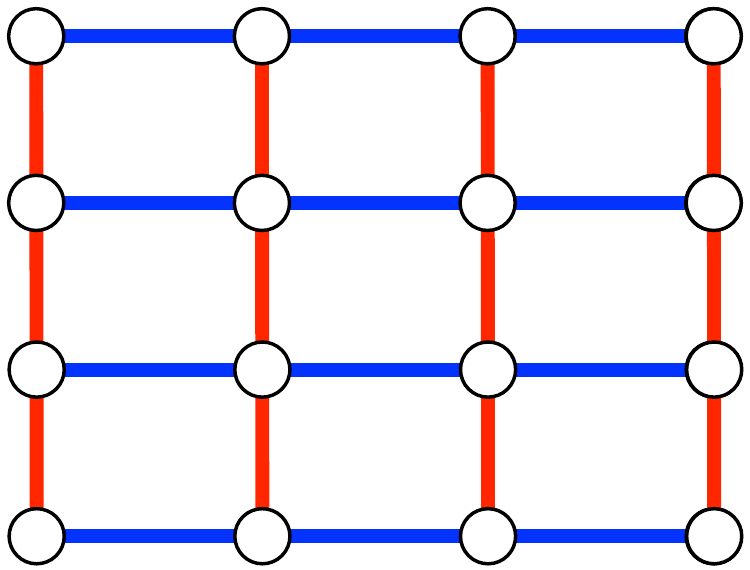} \hspace{10pt}
  \includegraphics[width=0.28\textwidth]{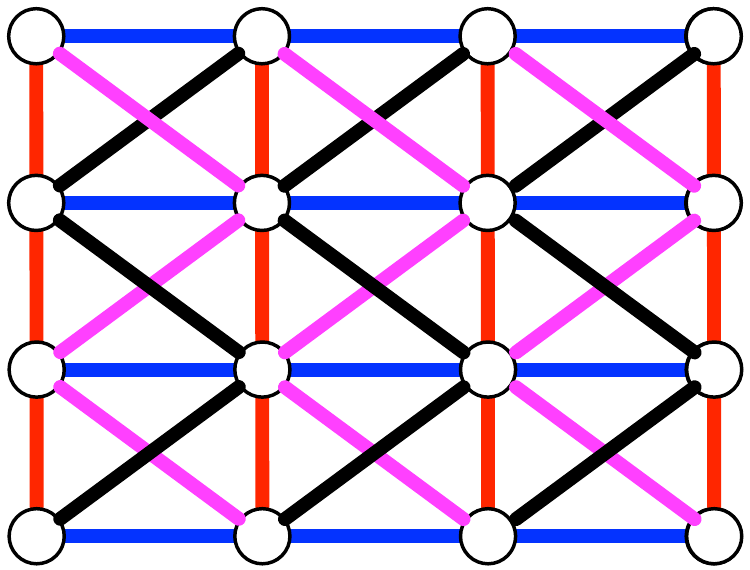} \hspace{10pt}
  \includegraphics[width=0.31\textwidth]{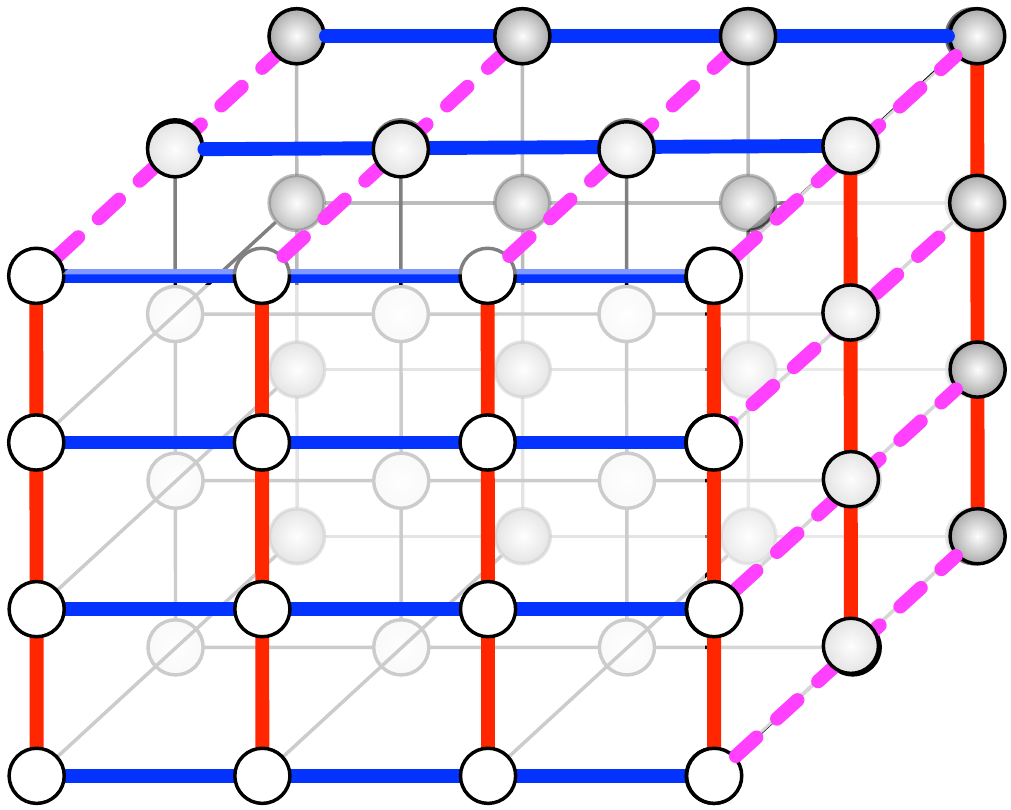}
  \caption{Grid graph structures and decompositions, indicated by different colors. Left: 2D, 4-connected, middle: 2D, 8-connected, right: 3D tensor, 6-connected.}
  \label{fig:grids2D}
\end{figure}

The most frequenly used 2D and 3D topologies in computer vision decompose straightforwardly in several ways~\cite{strandmark2010parallel,komodakis2011mrf,kolmogorov12,savchynskyy11}. 
In our experiments, we split the 2D 4-connected grid into vertical and horizontal lines (Figure~\ref{fig:grids2D} left), and the 2D 8-connected grid additionally into zig-zagged paths (Figure~\ref{fig:grids2D} middle) . Diagonal lines are also possible but less efficient with respect to memory access.

Even though the set of 1D lines of equal color in the figures may be considered one $f_j$, the lines are independent and may be addressed in parallel. Further parallelization (across colors) results from our splitting formulation.
Finally, any fast algorithm for solving a 2D or 3D TV (or graph cut) problem may be used as a subroutine when setting $f_j$ to be an entire 2D grid sheet or 3D tensor.

\subsection{Efficient 1D TV}
\label{sec:1DTV}
As chains are the building blocks of our decomposition, performance of the overall method is heavily influenced by the speed at which 1D-TV can be solved. Being a classic regularizer for image denoising, literature on solvers for different variations of 1D and 2D TV abounds, though only recently fast direct methods for chains have been proposed. A notable example is the dynamic programming method of Johnson~\cite{dpTV}, which guarantees linear complexity. Another outstanding method is that of Condat~\cite{fastTV}; it is based on a thorough analysis of the KKT conditions and manages to achieve faster running times in practice, despite a pathological quadratic cost worst-case. These TV solvers, however, only apply to chains with constant weights.

To permit varying weights we use a recent method of~\cite{barberoTV14} that obtains Condat's method through a taut-string viewpoint~\cite{grasmairTV07}, in a way that allows weights along the chain. Experiments~\cite{barberoTV14} indicate that this method shares the same performance as the original procedure, therefore rapidly solving TV chains in linear time in practice.

We also point out that the choice of the 1D-TV solver is independent of the overlying topologies and optimizers. This allows us to localize complexity to highly tuned TV chain solvers for the architecture under use (multicore, GPUs, etc.), thus providing overall modularity and adaption to the underlying hardware. In this paper, we use a general implementation for CPUs.

\paragraph{Message passing.}
Alternatively to the method that we have used, we may also use message passing techniques, which could be more efficient on certain architectures. These are directly adapted to solve the min-cut problem on a chain or a tree, not the total variation problem. However, it is known that by a sequence of at most $n$ min-cut problems, one may obtain the exact TV solution~\cite{hochbaum95}.

\subsection{Decomposed dual problems}
\label{sec:duals}
By their form (total variation or Lov\'asz extensions of submodular functions), the $f_i$ 
may be represented as a maximum of linear functions, that is,  $f_j(x) = \max_{y_j \in K_j}\ y_j^\top x$, for $K_j$ a certain polytope, $j \in \{1,\dots,r\}$. 
This form as well as the decomposability of the total variation may be used to obtain a decomposed dual problem for the continuous Problem \eqref{eq:C}. The dual splits in the same way as the primal, and admits parallel optimization algorithms~\cite{strandmark2010parallel,komodakis2011mrf,kolmogorov12,savchynskyy11}. It however has a non-smooth objective function that makes optimization harder.

We hence next describe two dual problems for the (TV) problem \cite{treesubmod}.

\paragraph{First dual problem.}
We use a standard reformulation for dual decomposition: we introduce a variable $\mathbf{x} = (x_1,\dots,x_r) \in \rb^{n} \times \cdots \times \rb^n$ composed of $r$ copies of the input variable $x$, with the constraint that $x_i = x$ for each $i \in \{1,\dots,r\}$ (see, e.g.,~\cite{boyd.admm}). We then add Lagrange multipliers $\lambda_i \in \rb^n$ for each of these constraints. Writing $f_j$ as a maximum of linear functions introduces a dual variable vector $y_j  \in K_j$. We collect those variables in a vector $\mathbf{y} = (y_1,\dots,y_r) \in \mathbf{K} = K_1 \times \cdots \times K_r$. Overall, we obtain the following:

\begin{align}\displaybreak[3]
  \nonumber
  &\min_{x \in \rb^n} \;\; \sum_{j=1}^r f_j(x)  + \frac{1}{2} \| w - x\|^2 \\
  \nonumber
  \displaybreak[3]
  &\quad =\; \max_{\boldsymbol{\lambda} \in \rb^{n \times r}, \ \mathbf{y} \in \mathbf{K}} \ \min_{x \in \rb^n, \ \mathbf{x} \in \rb^{n\times r} }
\sum_{j=1}^r x_j^\top y_j   + 
\frac{1}{2} \| w\|^2  
 - w^\top x + 
\frac{1}{2r } \sum_{j=1}^r \|  x_j\|^2  + \sum_{j=1}^r \lambda_j^\top ( x - x_j) \\ 
\nonumber
\displaybreak[3]
&\quad = \max_{\boldsymbol{\lambda} \in \mathbf{L}, \ \mathbf{y} \in \mathbf{K} } 
\frac{1}{2} \| w\|_2^2 - \frac{r}{2} \sum_{j=1}^r \| y_j  -  \lambda_j \|^2\\
\displaybreak[3]
\label{eq:dualdec}  &\quad = \max_{\boldsymbol{\lambda} \in \mathbf{L}, \ \mathbf{y} \in \mathbf{K} } 
\frac{1}{2} \| w\|_2^2 - \frac{r}{2} \| \mathbf{y} -  \boldsymbol{\lambda} \|^2, 
\end{align}
where $\mathbf{L}$ denotes the set of  $\boldsymbol{\lambda} \in \rb^{n \times r}$ such that
$\displaystyle \sum\nolimits_{j=1}^r \lambda_j = w$. We are thus faced with the problem of finding the closest point between two convex sets, which we explore in Sections~\ref{sec:AP} and~\ref{sec:AR}.

Note that if the function $f_j$ only depends on a subset of variables of $x$, then we may restrict the corresponding variable $\lambda_j$ to be zero on the complement of that subset in order to have faster convergence for the iterative methods presented below.

\paragraph{Second dual problem.}
Given $ \mathbf{y} \in \mathbf{K}$ in \eqref{eq:dualdec}, the optimal $\boldsymbol{\lambda} \in \mathbf{L}$ may be obtained in closed form:
\begin{equation}
\label{eq:dualA}
\lambda_j = \frac{w}{r} + y_j  - \frac{1}{r} \sum_{k=1}^r y_k.
\end{equation}
This leads to another dual problem, where the variables $\boldsymbol{\lambda} \in \mathbf{L}$ are maximized out:
\begin{equation}
  \label{eq:dualdec2} 
 \max_{ \mathbf{y} \in \mathbf{K} } \;\;\;
\frac{1}{2} \| w\|_2^2 - \frac{1}{2}  \big\|  \sum\nolimits_{j=1}^r y_j  - w \big\|^2.
\end{equation}
The problem above has \emph{separable constraints} $y_i \in K_j$, $j \in \{1,\dots,r\}$ and a smooth objective function. We will discuss optimization procedures in Section~\ref{sec:cyclic}.

\paragraph{Special case $r=2$.}
When the function $f$ is split into two functions, then the problem in  \eq{dualdec2} is equivalent to finding the distance between the convex set $K_1$ and the set $\{w - y_2 \mid \ y_2 \in K_2\}$, for which methods presented in Sections~\ref{sec:AP} and~\ref{sec:AR} may be used.

\begin{figure}
  \centering
  \begin{tabular}{@{\hspace{0pt}}c@{\hspace{0pt}}c@{\hspace{0pt}}c@{\hspace{0pt}}c}
    {\small AP, iter 1} &     {\small AP, sequence} &     {\small AAR, iter 1} &     {\small AAR, sequence} \\
    \includegraphics[width=0.25\textwidth]{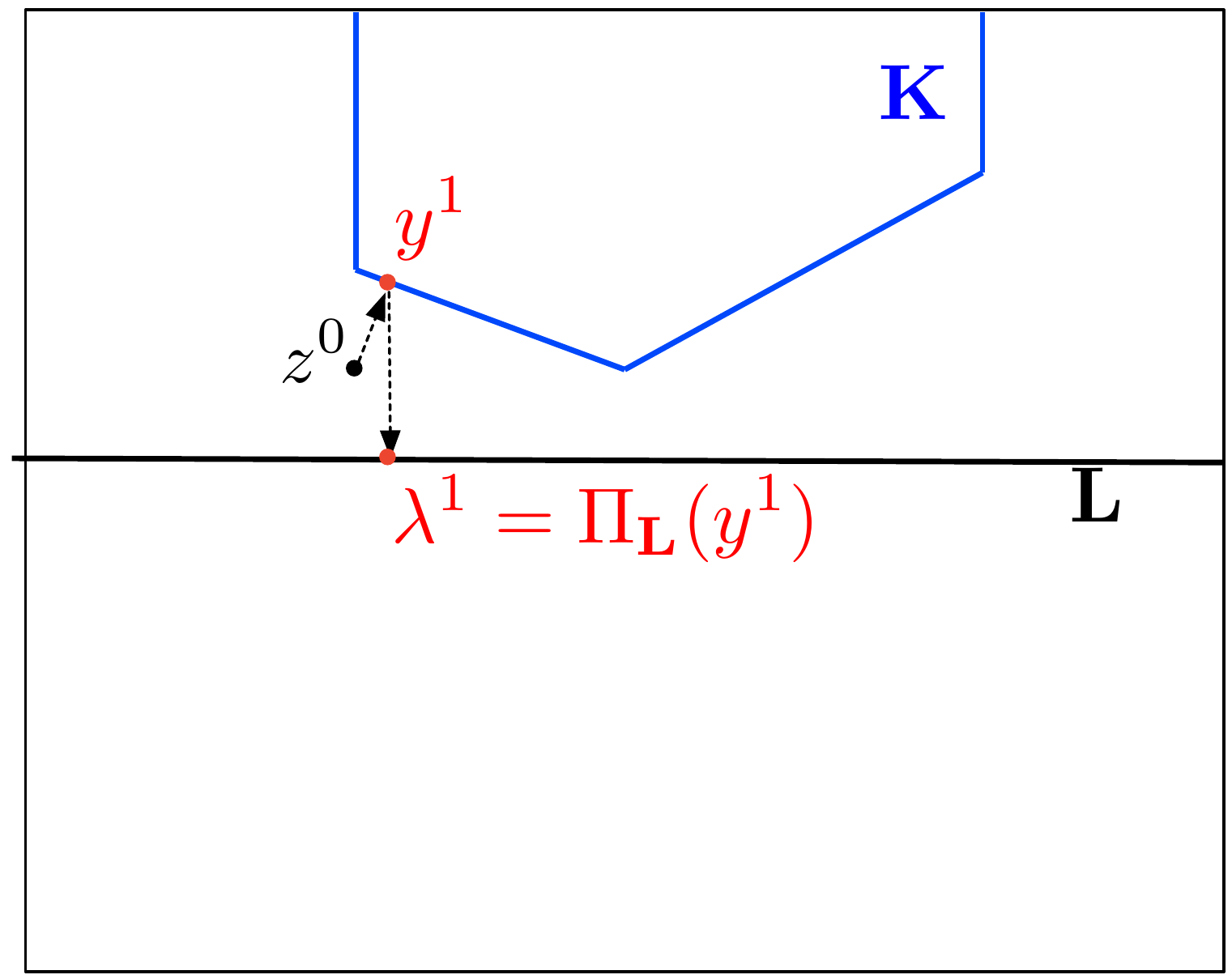} &
    \includegraphics[width=0.25\textwidth]{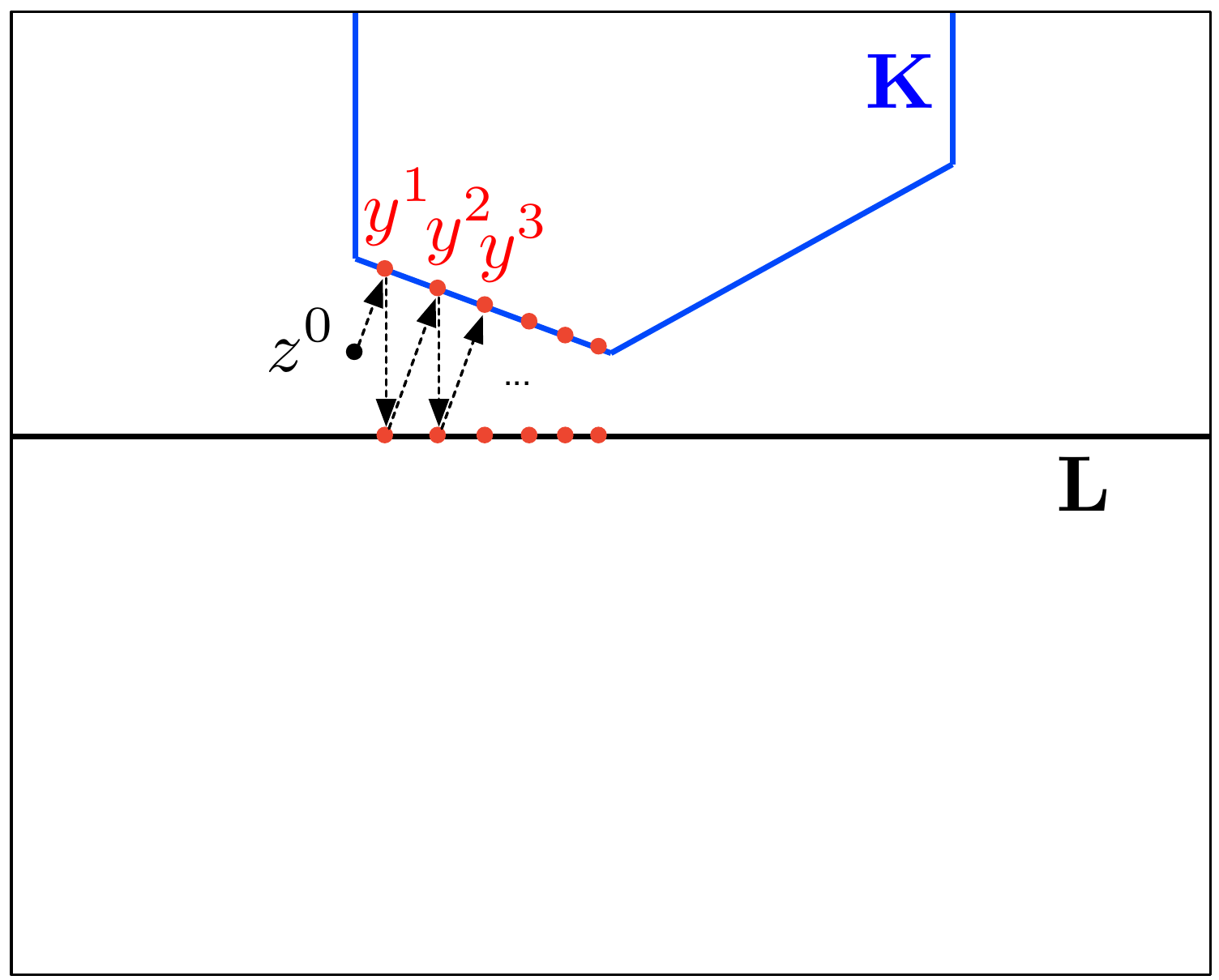} &
    \includegraphics[width=0.25\textwidth]{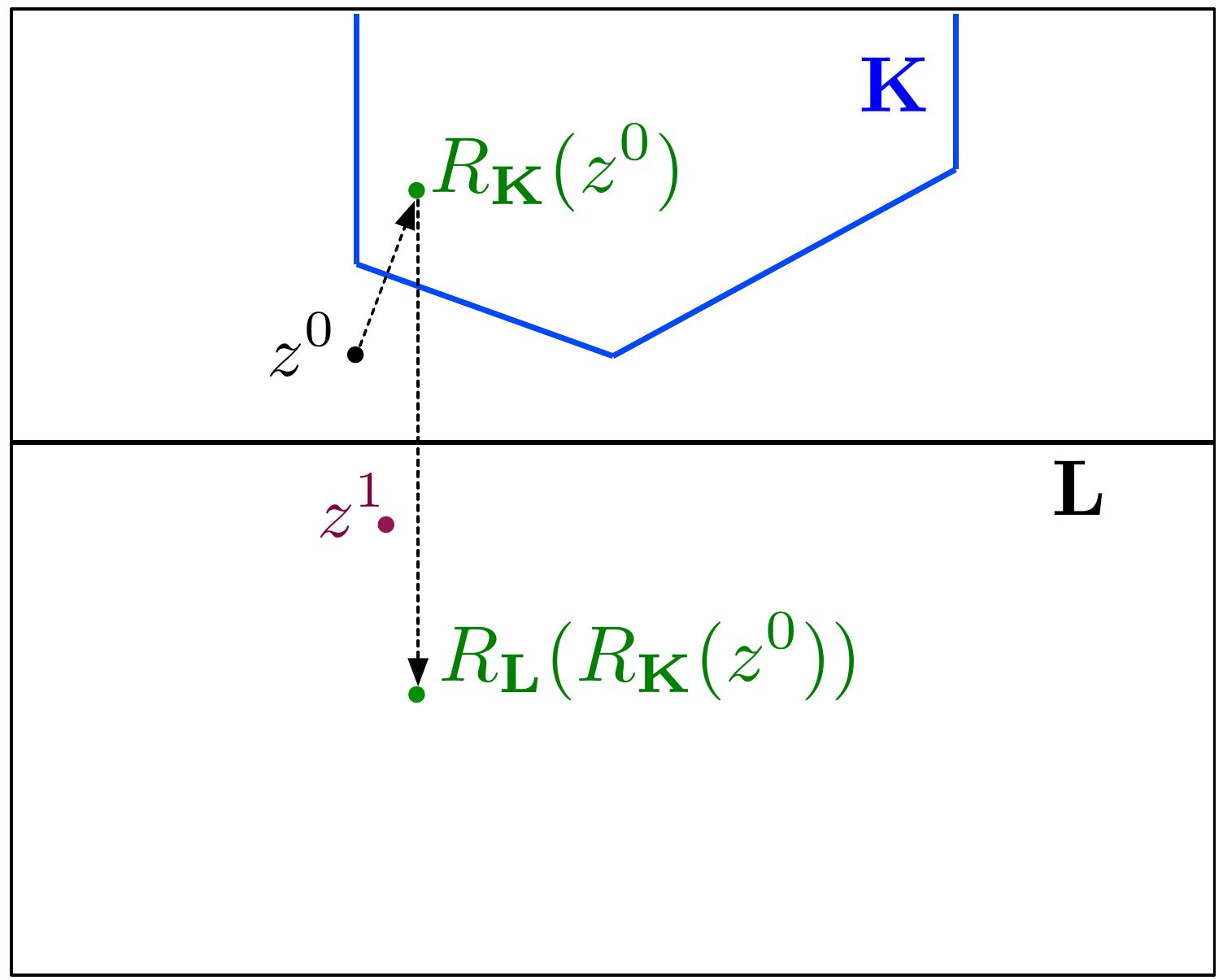} &
    \includegraphics[width=0.25\textwidth]{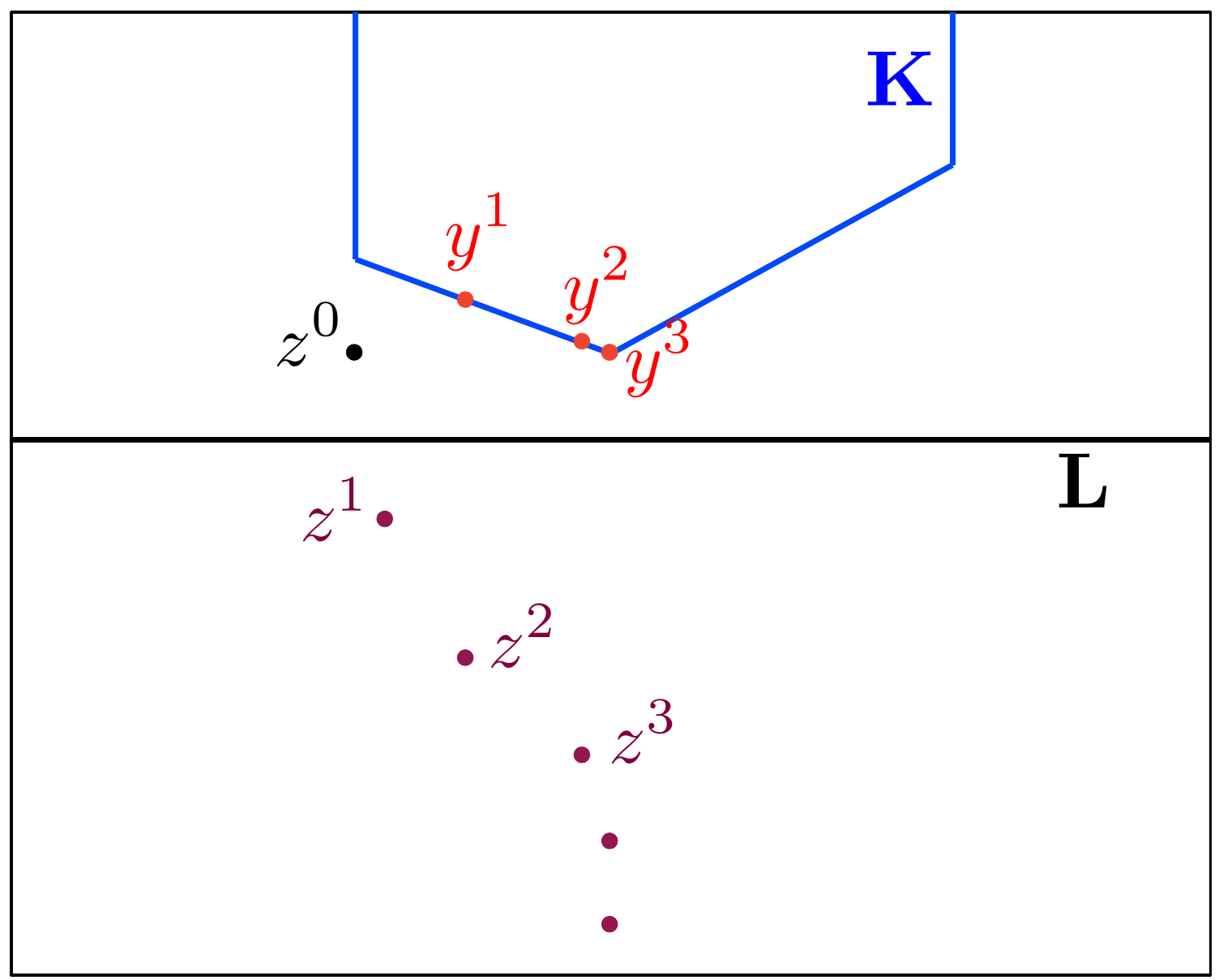}
  \end{tabular}
  \caption{Illustration of alternating projections (AP) and averaged alternating reflections (AAR) for Problem~(\ref{eq:dualdec}), the problem of finding the closest points between a polytope $\mathbf{K}$ and a subspace $\mathbf{L}$. The iterations start at the black point $z^0$. AP alternatingly projects onto $\mathbf{K}$ and then $\mathbf{L}$. AAR reflects at $\mathbf{K}$ and then at $\mathbf{L}$; the next iterate $z^{t+1}$ (magenta) is the midpoint between $z^t$ and $R_{\mathbf{L}}R_{\mathbf{K}}(z^t)$. For AAR, the sequence $z^t$ diverges, but the projected shadow sequence of $y^t = \Pi_{\mathbf{K}}(z^t)$ converges (red). Here, AAR takes larger steps than AP and hence converges more quickly.
  }
  \label{fig:ap_vs_aar}
\end{figure}

\section{Optimization for decomposable problems}
Next, we describe optimization procedures that exploit the decomposable structure of the dual problems~\eqref{eq:dualdec} and~\eqref{eq:dualdec2}, where in particular the Cartesian decomposition $\mathbf{K} = K_1\times \cdots \times K_r$ of the constraint sets plays an important role. In particular, we exploit that the projection onto $\mathbf{K}$ consists of $r$ independent projections onto the sets $K_i$. By the above derivations, each of those projections can be done quickly via a TV subroutine for each $f_i$.

\subsection{Cyclic projections}
\label{sec:cyclic}
The first method we consider for problem~\eqref{eq:dualdec2} is \emph{block coordinate descent} (BCD), a classic method~\cite{bertsekas1999nonlinear} that has recently witnessed a huge resurgence of interest in large-scale optimization~\cite{richtarik,beck}. Since the cost function is a separable quadratic, BCD assumes a form that is more commonly known as \emph{cyclic projections} (more precisely, this is so if we go through the constraint blocks in a cyclic order). Specifically, we update coordinate blocks $i=1,\ldots,r$ as follows:
\begin{align*}
  y_i \gets \argmax_{z \in K_i}\ -\frac12\bigl\|z -w +\sum_{j\neq i}y_i\bigr\|^2=
  \Pi_{K_i}\bigl(w-\sum_{j\neq i}y_i\bigr),
\end{align*}
where $\Pi_{K_i}$ denotes orthogonal projection onto set $K_i$. This projection is solved by solving a (fast) TV problem with $f_i$. Notice that the variable $y_i$ is overwritten after the update, so that when updating $y_{i+1}$, the latest values $y_1,\ldots,y_i$ are used in the projection.

In addition to cyclic projections, one could solve the smooth dual~\eqref{eq:dualdec2} using a gradient-based method like FISTA~\cite{fista}. Such a method is also easy to implement because the Lipschitz constant of the gradient is easily seen to be $r$ and the required projections decompose due to the  structure of $\mathbf{K}$.

\subsection{Alternating projections in product space}
\label{sec:AP}
The method of cyclic projections offers a practical choice. However, it is inherently serial. To solve the problem in parallel, the first dual formulation~\eqref{eq:dualdec} turns out to be more suited (note that this provides a second source of parallelization, beyond the fact that each polytope $K_j$ is itself a product of polytopes corresponding to individual lines).

The key idea is to  exploit the ``product space'' $K_1\times\cdots\times K_r$. Since $w$ is constant, as previously mentioned, \eqref{eq:dualdec} is nothing but the problem of finding the closest point between two convex sets~\cite{treesubmod}. Applying BCD, except this time with just two coordinate blocks, we obtain the classic \emph{alternating projections} (AP) (cast in a product space setting), which performs for $k=0,1,\ldots,$ the iteration:
\begin{align*}
  \vy^{k+1} &\gets \argmax_{\vy \in \mathbf{K}}\ -\frac{r}{2}\norm{\vy - \boldsymbol{\lambda}^k}^2 = \Pi_{\mathbf{K}}(\boldsymbol{\lambda}^k),\\
  \boldsymbol{\lambda}^{k+1} &\gets\argmax_{\boldsymbol{\lambda} \in \mathbf{L}}\ -\frac{r}{2}\norm{\boldsymbol{\lambda}-\vy^{k+1}} = \Pi_{\mathbf{L}}(\vy^{k+1}).
\end{align*}
The key point here is that the projection $\Pi_{\mathbf{K}}$ decomposes
\begin{equation*}
  \Pi_{\mathbf{K}}(\boldsymbol{\lambda}) = (\Pi_{K_1}(\lambda_1), \ldots, \Pi_{K_r}(\lambda_r)),
\end{equation*}
so that each of the coordinate blocks may be computed in parallel (our implementation exploits this fact), while the projection $\Pi_{\mathbf{L}}$ is merely an averaging step detailed in \eqref{eq:dualA}.

\subsection{Alternating reflections in product space}
\label{sec:AR}
The recent work~\cite{treesubmod} provided strong experimental evidence that for projection problems of the form~\eqref{eq:dualdec}, AP is often outperformed by a more refined method of~\cite{bauschke2004finding}, namely, \emph{averaged alternating reflections (AAR)}. Here, instead of alternating between the projection operations $\Pi_{\mathbf{K}}$ and $\Pi_{\mathbf{L}}$, one uses \emph{reflection operators}
\begin{equation}
  \label{eq:2}
  R_{\mathbf{K}} := 2\Pi_{\mathbf{K}} - I,\quad R_{\mathbf{L}} := 2\Pi_{\mathbf{L}} - I,
\end{equation}
while averaging them to ensure firm nonexpansivity, a property that greatly simplifies convergence analysis~\cite{bauschke2004finding}. To apply the AAR method, one first introduces the auxiliary vector $\mathbf{z}$, which represents $\mathbf{y}-\boldsymbol{\lambda}$. Then, AAR takes the form
\begin{equation}
\label{eq:3}
  \mathbf{z}^{k+1} = \half(R_{\mathbf{L}}R_{\mathbf{K}} + I)\mathbf{z}^k.
\end{equation}
However since usually $\mathbf{K} \cap \mathbf{L} = \emptyset$, the sequence $(\mathbf{z}^k)$ generated by~\eqref{eq:3} diverges to infinity! The remarkable fact is that from this diverging sequence, we can extract a solution by maintaining a ``shadow sequence'' $\vy^{k} \equiv \Pi_{\mathbf{K}}(\mathbf{z}^k)$. See Figure~\ref{fig:ap_vs_aar} for an illustration, and Theorem 3.13 in~\cite{bauschke2004finding}  for a proof of convergence.

\subsection{Extensions}
\label{sec:extensions}
Above, we outlined flexible, parallelizable convex optimization algorithms for energy minimization with pairwise submodular potentials. These algorithms straightforwardly generalize from binary labels to the multi-label case, to submodular higher-order potentials, and to related problems. The reasons are two-fold: (1) the above algorithms solve a minimum cut problem, and any methodological machinary that builds on graph cuts as a subroutine will work with the above algorithms too; (2) the decomposition theory and tightness of the relaxations hold generically for submodular functions, not only graph cuts.

For multi-label energy minimization, one may use move-making algorithms \cite{boykov2001fast} that reduce the multi-label problem to a series of binary submodular energy minimization problems. The methods above solve those binary problems. For combinatorial algorithms, it has proved useful to reuse existing solutions and data structures \cite{kohli05dynamic}. ``Warm-starting'' is possible for the convex case too: we simply use the $y_i$ vectors of the previous problem to initialize the new problem. If the geometry of the polytopes $K$ has not changed too much, this can save many iterations (see \myfig{coreswarm}(b)).

Second, the convex approach directly generalizes to submodular potentials that involve more than two nodes at a time (following \cite{treesubmod}); such potentials include \cite{chambolle2009total,stobbe11,kohli09P3,kohli09robust,hein13}. Many of those potentials correspond to sufficiently simple submodular functions, often with small support, such that the relaxation  (the equivalent to total variation for graph cuts) can be solved fast.
Moreover, the same methods may even generalize to be used with roof duality~\cite{rother2007optimizing}. 

Finally, since the above methods also solve the parametric version of the discrete problem (by thresholding the solution of \eq{TV} at different levels) as a byproduct, they are also applicable to the numerous applications of parametric graph cuts \cite{kolmogorov07,hochbaum13}.

\section{Implementation details}
The algorithms are inherently parallel by design as each projection/reflection onto a chain graph is independent of the other. Our implementation assumes decomposed functions and the decomposition depends on the problem at hand. In \mysec{2D3D}, we described some possible decompositions of grid-like graph structures on 2D and 3D graphs. However, this extends to many other decompositions. Empirically and theoretically \cite{nishihara14}, longer connected structures lead to faster convergence than decomposition e.g. into single edges.

\subsection{Parallelization}
We use the efficient 1D-TV implementation of \cite{barberoTV14} to solve the projection/reflection on chains in parallel. Our implementation is in C++ and uses OpenMP; it ensures that the memory access pattern across threads is streamlined, since bad memory access patterns can lead to considerable slowdowns. While the 1D-TV solver is not optimized for GPUs, as explained in \mysec{1DTV}, it can be replaced by message passing based subroutines, which are inherently parallel and also friendly to GPU architectures. 

\subsection{Memory footprint}
In our implementation every decomposable function must maintain states of dual variables for each node in the graph. Thus, the memory requirement of our methods increases bilinearly in the number of decomposed functions and the number of nodes. Our experiments suggest that the projection-based algorithms require less memory than standard combinatorial algorithms---see Table~\ref{table:memory}. Unlike the 32 bit integers used in many other implementations, we use (64 bit) double precision numbers. Reducing those to 32 bit would reduce the memory requirements of the projection methods even further. 

\begin{table}[htbp]
\begin{center}
\begin{tabular}{ccccc}
\toprule
Algorithm   & AAR    & BK~\cite{boykov04}& IBFS~\cite{ibfs} & HPF~\cite{Chandran:2009}  \\
\hline
Memory(GB) &26.82 & 42.83 & 44.16& 55.46 \\
\bottomrule
\hline
\end{tabular}
\end{center}
\caption{Memory footprint for {\em Abdomen} dataset ($512\times512\times551$)}
\label{table:memory}
\end{table}

\subsection{Running time}
With the TV subroutine we use, each projection/reflection step scales in the worst case quadratically in the length of the chain (the chain length is
typically equal to $\sqrt{n}$ or $\sqrt[3]{n}$, where $n$ is the total number of nodes in the 2D or 3D graphs), but is in practice usually linear~\cite{barberoTV14}. In fact, it did not scale quadratically for any of our data. Hence, empirically, the cost of each iteration grows bilinearly with the number $f$ of functions $f_i$, and with the number of nodes in the graph.

\section{Experiments}

\begin{figure}
  \centering
  \begin{tabular}{cc}
  \includegraphics[width=0.3\textwidth]{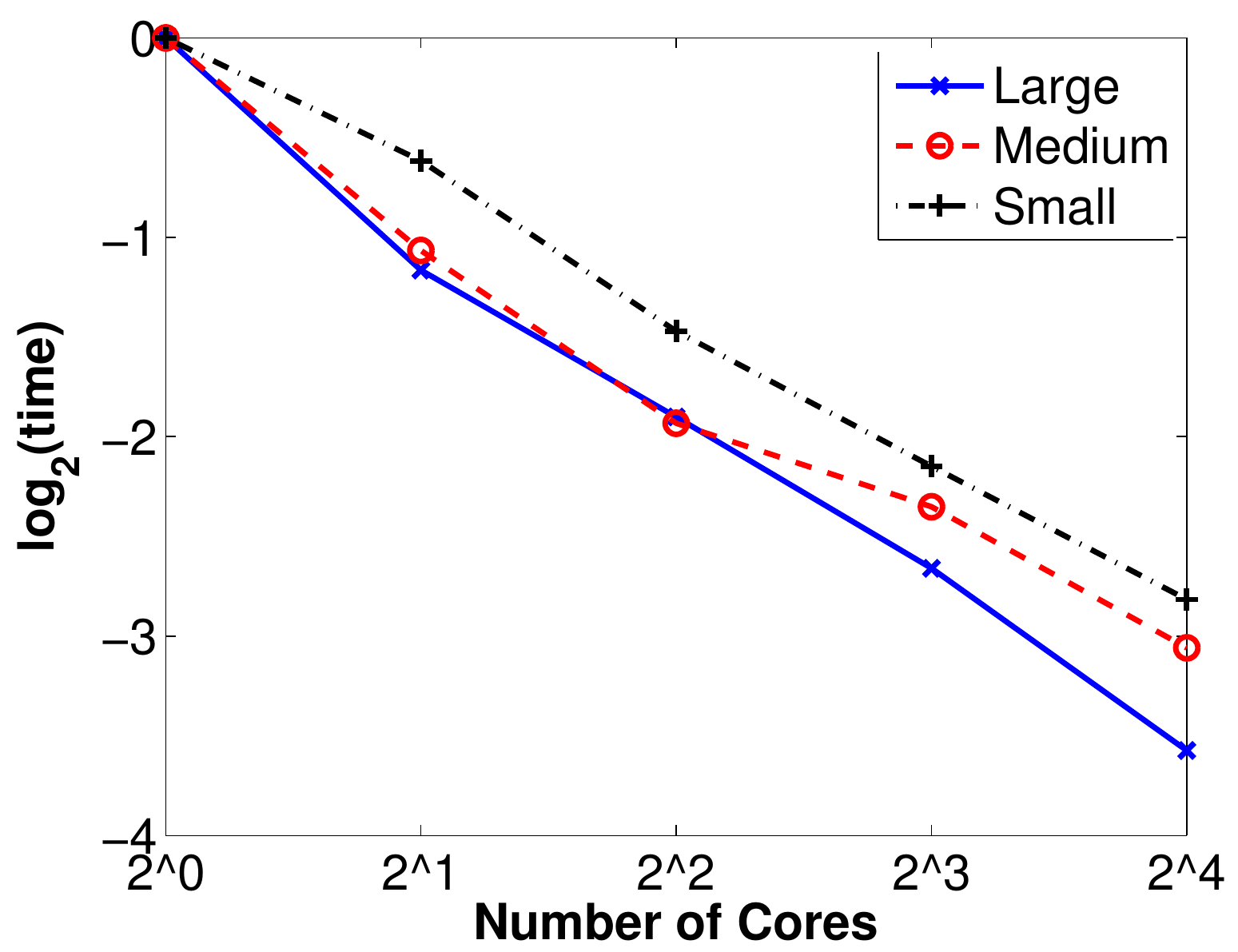} &
  \includegraphics[width=0.3\textwidth]{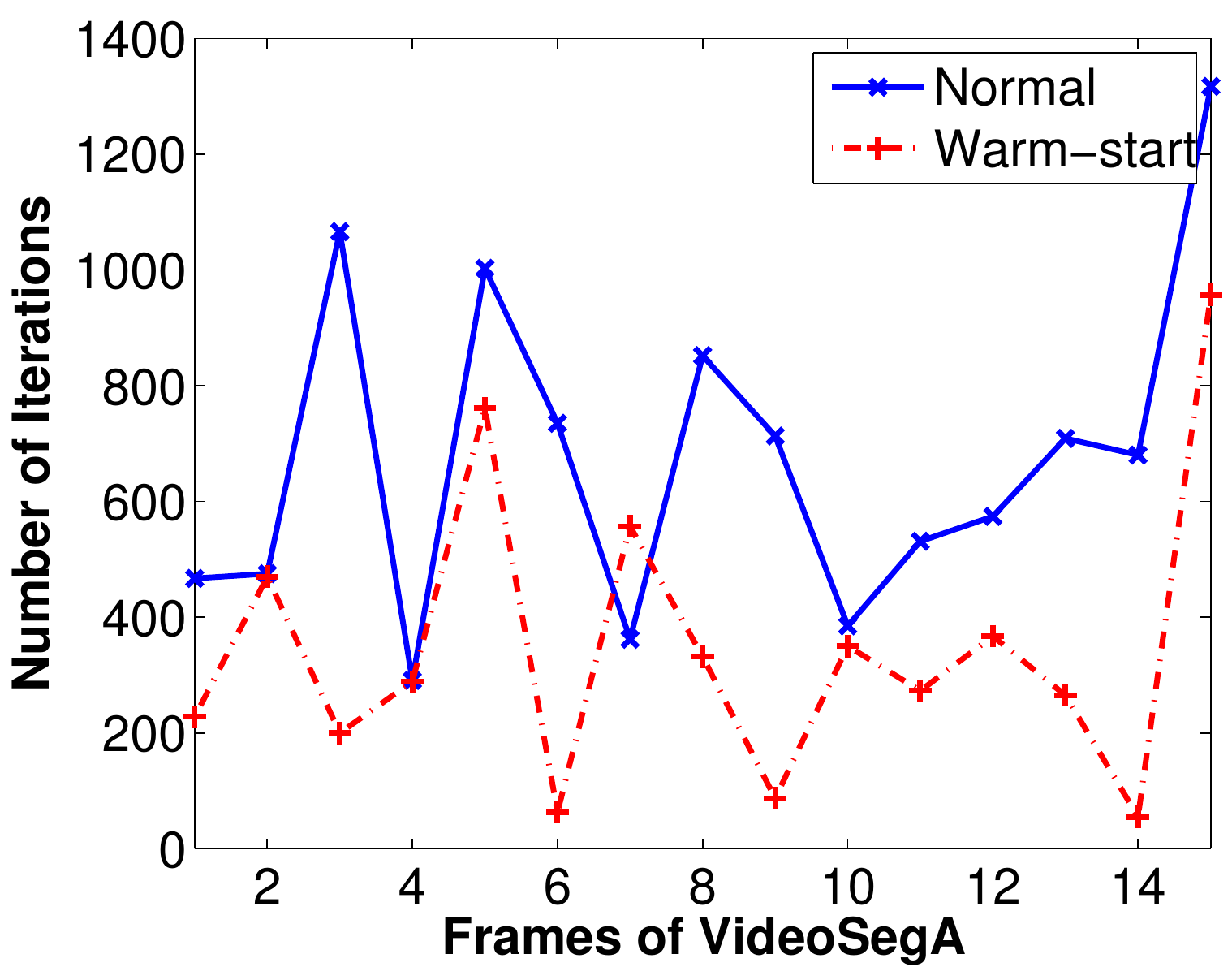} \\
   (a) & (b) \\
   \end{tabular}
   \caption{Speedup of the AAR algorithm with parallelization and warm starts. (a) Normalized scale of performance with increasing number of cores on Bunny datasets~\cite{maxflowDataset} of different resolutions. (b) Number of iterations taken by each from of a video with "normal" initialization and "warm start" from the dual variables of the previous frame.}
  \label{fig:coreswarm}
\end{figure}

Our experiments study the performance of the projection algorithms on 2D and 3D-maxflow datasets~\cite{maxflowDataset}, exploiting in particular the parallel nature of the algorithms. We compare the  algorithms to standard, popular maxflow implementations such as BK~\cite{boykov04}, IBFS~\cite{ibfs}, and HPF~\cite{Chandran:2009}. For other algorithms~\cite{strandmark2010parallel,komodakis2011mrf}, we have not been able to find implementations that were easily portable to 3D datasets.

Table~\ref{table:timeComparison} shows running time and the number of iterations for the projection algorithms and others on a multicore machine\footnote{20 core, Intel(R) Xeon(R) CPU E5-2670 v2 @ 2.50GHz with 100Gigabytes of memory. We only use up to 16 cores of the machine to ensure accurate timings.}. The timings are recorded using the {\em gettimeofday} command. All the timings are for the optimization phase only and exclude data I/O (which is common to all methods).

\subsection{2D problems}
As a 2D example, we use the tsukuba data~\cite{maxflowDataset}, a multi-label task on 2D images corresponding to 4-neighborhood grids. To cope with multiple labels, we use alpha expansion \cite{boykov2001fast}. Notably, the decomposition for 2D 4-connected grids uses only two functions (vertical and horizontal), and therefore corresponds to the special dual for $r=2$ in Section~\ref{sec:duals}. For this formulation, AAR converges remarkably faster than other iterative algorithms, and even outperforms the combinatorial methods. The time comparisons at the bottom of Table~\ref{table:timeComparison} show that in general, the running times of our methods are comparable to standard combinatorial algorithms on images of $384 \times 288$ size. 

\paragraph{Warm starts.}
\myfig{coreswarm}(b) shows the number of iterations required for the algorithm to converge on each frame of size $480 \times 360$ of the VideoSegA~\cite{maxflowDataset} dataset. These are the consecutive frames of a video, which are 2D images with 8-neighborhood grids (see Figure~\ref{fig:grids2D}). We use the dual variables at the convergence of the previous frame to warm-start the projection/reflection process. This makes the method converge to the optimal solution substantially quicker than with other initializations.

\begin{sidewaystable}
\begin{center}
\begin{tabular}{llrrrrrrrrlrrrrr}
\toprule
\multirow{2}{*}{Dataset}     &  & \multicolumn{8}{c}{Time in seconds}                                       & & \multicolumn{5}{c}{Iterations}                         \\
\cline{3-10}\cline{12-16}       
                             &  & AAR               & AAR               & AAR-JD           & AAR-JD            &  AAR    & BK     & IBFS  & HPF    & &   AAR& AAR           &  AP   &   BCD &FISTA            \\
                             &  &\small{($<10\%$)}  &\small{($<2\%$)}   &\small{($<0.1$)}  &\small{($<0.02$)}  &         &        &       &        & &      & \small{(0.1x)}&       &       &                 \\
\cline{1-1}\cline{3-10}\cline{12-16}                      
(3D) bone-100                     &  &   4.2             &  7.4              &  3.9             &  10.9             &    14.4 &  8.5   & 6.3   &1.0     & &  105 &  73           &  846  &  146  &   422            \\
(3D) bone-10                      &  &   4.5             &  7.4              &  4.7             &  9.6              &    18.5 &  5.6   & 3.4   &0.9     & &  134 &  25          & 1183  &  206  &   592            \\
\cline{1-1}\cline{3-10}\cline{12-16}                     
(3D) bone\_x-100                  &  &   0.08            &  0.08             &  0.08            &  0.08             &     3.90&  2.42  & 1.02  &1.08    & &   45 &  34           &  42   &   9   &   25             \\
(3D) bone\_x-10                   &  &   0.09            &  0.09             &  0.09            &  0.09             &     3.70&  1.86  & 0.93  &0.75    & &   45 &   23          &  44   &   10  &   26             \\
\cline{1-1}\cline{3-10}\cline{12-16}                    
(3D) bone\_xy-100                 &  &   0.01            &  0.01             &  0.02            &  0.02             &     1.25&  0.79  & 0.68  &0.40    & &   44 &  36           &  26   &    6  &   16             \\
(3D) bone\_xy-10                  &  &   0.01            &  0.01             &  0.03            &  0.03             &     1.18&  0.64  & 0.53  &0.29    & &   36 &    21          &  27   &    7  &   17             \\
\cline{1-1}\cline{3-10}\cline{12-16}                   
(3D) bone\_xyz-100                &  &   0.39            &  0.54             &  0.51            &  0.51             &     0.91&  0.47  & 0.32  &0.16    & &   57 &   43          & 185   &  36   &  98              \\
(3D) bone\_xyz-10                 &  &   0.34            &  0.57             &  0.46            &  0.48             &     0.82&  0.36  & 0.33  &0.11    & &   57 &   21          & 185   &  37   &  98              \\
\cline{1-1}\cline{3-10}\cline{12-16}                  
(3D) bone\_xyz\_x-100             &  &   0.42            &  0.43             &  0.42            &  0.43             &     0.53&  0.14  & 0.09  &0.06    & &   53 &    48         & 618   &  88   & 256              \\
(3D) bone\_xyz\_x-10              &  &   0.43            &  0.47             &  0.39            &  0.39             &     0.48&  0.12  & 0.03  &0.05    & &   50 &   23          & 615   &  97   & 290              \\
\cline{1-1}\cline{3-10}\cline{12-16}                 
(3D) bone\_xyz\_xy-c100           &  &   0.24            &  0.24             &  0.21            &  0.21             &     0.26&  0.05  & 0.03  &0.03    & &   29 &   28          & 117   &  25   &  63              \\
(3D) bone\_xyz\_xy-c10            &  &   0.18            &  0.22             &  0.18            &  0.22             &     0.24&  0.04  & 0.02  &0.02    & &   30 &   23          & 120   &  25   &  64              \\
\cline{1-1}\cline{3-10}\cline{12-16}                       
(3D) babyface-100                 &  &   3.6             &  8.2              &  9.2             &  32               &    33.5 & 25.7   & 12.3  &9.4     & &  509 &   346         & 873   &  550  & 1360            \\
(3D) babyface-10                  &  &   6.2             &  9.4              &  7.5             &  13.9             &    35.2 & 14.3   & 7.9   &7.6     & &  543 &   223          & 793   &  420  & 1162            \\
\cline{1-1}\cline{3-10}\cline{12-16}                
(3D) bunny-lrg                    &  &  16.7             & 28.3              & 1.28             & 1.28              &   186.6 &  9.5   & 6.3   &41.2    & &  145 &   52          & 796   &  133  & 406              \\
(3D) bunny-med                    &  &   1.72            &  2.72             & 0.14             & 0.14              &     7.47&  1.07  & 1.27  &2.27    & &   52 &   25          & 94    &  17   &  52              \\
(3D) bunny-sml                    &  &   0.12            &  0.19             &  0.11            &  0.11             &     0.38&  0.11  & 0.17  &0.23    & &   35 &   16          & 111   &  18   &  58              \\
\cline{1-1}\cline{3-10}\cline{12-16}               
(3D) liver-100                    &  &  10.3             & 15.0              & 4.88             & 4.88              &     38.5&  7.3   &  4.7  &4.7     & &  654 &   503         & 1682  &  1444 & 2873             \\
(3D) liver-10                     &  &  10.4             & 15.8              & 5.06             & 5.06              &     33.1&  3.4   &  3.2  &3.3     & &  523 &    407          & 1586  &  1290 & 2754             \\
\cline{1-1}\cline{3-10}\cline{12-16}                         
(3D) abdomen\_long                &  & 525               & 701               & 441              & 1445              & 1445    & 212    & 110   & 68     & &  468 &  349          & 2532  &   939 & 1432            \\
(3D) abdomen\_short               &  & 578               & 772               & 468              & 1540              & 1593    & 119    &  60   & 29     & &  485 &   231          & 2373  &   953 & 1428            \\
\cline{1-1}\cline{3-10}\cline{12-16}                        
(3D) adhead-100                   &  &   9.5             & 27.2              & 0.2             & 8.5                &   42.2  &  10.1  &  8.1  &13.6    & &  208 &   176         & 453   &  104  & 148             \\
(3D) adhead-10                    &  &   9.1             & 25.4              & 0.2             & 4.5                &   42.3  &   6.3  &  8.4  &10.5    & &  208 &   105         & 395   &  111  & 176             \\
\cline{1-1}\cline{3-10}\cline{12-16}                        
(2D) BVZ-tsukuba                  &  &   0.15            &  0.18             & 0.02            & 0.01               &    0.21 &  0.31 & 0.20   & 0.24   & &   30 &    25         &  110  &   79  &  50              \\
\bottomrule
\end{tabular}
\caption{Performance comparison of AAR with BK~\cite{boykov04}, IBFS~\cite{ibfs}, and HPF~\cite{ibfs} on 3D datasets with 6 connectivity. AAR($< p\%$) denotes the time taken for the algorithm to find a cut whose difference to the optimal cut is $p\%$ of the difference between the cut in the first iteration and the optimal cut. AAR-JD($<p$) denotes time taken by the algorithm to reduce Jaccard Distance to $p$.
AAR(0.1x) is the number of iterations taken by AAR after scaling the pairwise weights by $0.1$. } 
\label{table:timeComparison}
\end{center}
\end{sidewaystable}

\subsection{3D problems}
On the 3D data, the running times of the algorithms differs more widely. 
In particular, the number of iterations for the algorithms to converge appears to depend on two important characteristics: (i) number of nodes in the graph (dimensionality) (ii) the edge weights in the graph (weights of the TV term). The latter affects the size of the polytopes $K$, i.e., the diameter of the domain of the dual problem, a parameter that commonly influences the convergence of convex optimization methods (see also \cite{nishihara14}).

\paragraph{Effect of edge weights.} Table~\ref{table:timeComparison} shows results for larger edge weights, as well as weights scaled by a factor of $0.1$. The iterative methods become faster with smaller weights, while the combinatorial methods robustly perform well with large weights too. On many instances, AAR converges faster than the cyclic BCD, while on others BCD is faster. 

\paragraph{Approximate solutions.} Since we can obtain a feasible solution (discrete cut) from any iterate of the projection methods by thresholding the continuous vector, Table~\ref{table:timeComparison} also shows the time taken to obtain an approximate solution with limited error (10\% and 2\%, measured by Jaccard distance). The results suggest that, while a complete dual certificate of convergence for the discrete problem takes a bit longer, a reasonable approximate solution can be obtained fairly quickly. 

\paragraph{Parallel speedup.} \myfig{coreswarm}-(a) shows the speedup of AAR achievable with an increasing number of cores. This figure reports the running time on the Bunny dataset (3D) with different resolutions: \emph{Large} ($401\times396\times312$), \emph{Medium} ($202\times199\times157$), and \emph{Small} ($102\times100\times79$). It is evident that more cores can improve the performance of the algorithm considerably. When using GPUs, it is important to consider their limited video memory, and hence the algorithms need to have a low memory footprint to perform well.

\paragraph{Memory.} Apart from running time, we also investigate the memory footprint of the algorithms. Table~\ref{table:memory} shows the memory footprint of all algorithms on the {\em Abdomen} data~\cite{maxflowDataset}, which is $512\times512\times551$. AAR uses considerably less memory than the standard algorithms.

\section{Conclusion and Future work}
\label{sec:conc}
We have proposed parallel iterative algorithms for binary energy minimization problems. The algorithms rely on a fast projection subroutine. For binary submodular potentials (graph cuts), this subroutine is simply a total variation problem, which can be efficiently solved on sub-graphs with special structure. In other examples, these subroutines could be fast algorithms for solving cuts on arbitrary subgraphs, or for simpler submodular energies. Hence, while the experiments here concentrate on cuts and decompositions into line graphs, the same methods apply to decompositions into 2D sheets, 3D cubes or any other subgraphs, and to sums of simple higher-order potentials.

We observed that the iterative methods perform similarly to combinatorial methods on 2D grid graphs, and require less memory than other, popular implementations of maximum flow algorithms.
The tradeoffs betweem convex and combinatorial methods illustrated here have some interesting implications, and suggest a wider study of integrating combinatorial and convex methods via different decompositions. For example, instead of TV oracles for line graphs, one may use oracles for larger specialized subgraphs. These oracles could use algorithms such as BK, HPF or IBFS, since the projection and TV oracle can be solved by parametric graph cuts. A 3D tensor is easily decomposed into two components: grids and lines. Those more complex subroutines can still be invoked in parallel using AAR. Thus, one can combine convex and combinatorial methods to greatly benefit from the strengths of both. 

The convex algorithms admit stochastic variants too \cite{Ene2015}. However, the decompositions used in the experiments here (Figure~\ref{fig:grids2D}) only use a decomposition into 2--4 functions.
Finally, given the remarkably improved behavior of iterative methods for smaller weights, it is of great empirical interest to study algorithms that can use homotopies or continuation techniques to start by solving with lower weights and use the ensuing solutions to speed up the medium and high weight regimes.

\bibliography{super_tree}

\end{document}